\documentclass[a4paper]{article}
\usepackage{latexsym}
\usepackage{graphicx}
\usepackage{amsfonts,amssymb,amsmath}
\usepackage{hyperref}
\usepackage{xcolor}
\usepackage{multirow}

\newcommand{\abele}{{\scshape abele}}
\newcommand{\lime}{{\scshape lime}}
\newcommand{\lore}{{\scshape lore}}

\newcommand{\shap}{{\scshape shap}}

\def\BibTeX{{\rm B\kern-.05em{\sc i\kern-.025em b}\kern-.08em T\kern-.1667em\lower.7ex\hbox{E}\kern-.125emX}}

\providecommand{\keywords}[1]
{
  \small	
  \textbf{\textit{Keywords---}} #1
}

\title{Explainable Deep Image Classifiers for Skin Lesion Diagnosis}

\begin{document}

\maketitle

\author{\center{Carlo Metta, Andrea Beretta, Riccardo Guidotti, Yuan Yin} \\ Patrick Gallinari, Salvatore Rinzivillo, Fosca Giannotti
\footnote{
    Carlo Metta, Andrea Beretta, and Salvatore Rinzivillo are with ISTI-CNR, Pisa, Italy (e-mail: carlo.metta@isti.cnr.it, andrea.beretta@isti.cnr.it, salvatore.rinzivillo@isti.cnr.it).
    Riccardo Guidotti is with University of Pisa, Italy (e-mail: riccardo.guidotti@unipi.it).
    Yuan Yin is with Sorbonne Université, Patrick Gallinari is with Sorbonne Université and Criteo AI Lab, Paris (e-mail: yuan.yin@sorbonne-universite.fr, patrick.gallinari@sorbonne-universite.fr),
    Fosca Giannotti is with Scuola Normale Superiore of Pisa, Italy (e-mail: fosca.giannotti@sns.it).
    }

\begin{abstract}
A key issue in critical contexts such as medical diagnosis is the interpretability of the deep learning models adopted in decision-making systems.
Research in eXplainable Artificial Intelligence (XAI) is trying to solve this issue.
However, often XAI approaches are only tested on generalist classifier and do not represent realistic problems such as those of medical diagnosis.
In this paper, we analyze a case study on skin lesion images where we customize an existing XAI approach for explaining a deep learning model able to recognize different types of skin lesions.
The explanation is formed by synthetic exemplar and counter-exemplar images of skin lesion and offers the practitioner a way to highlight the crucial traits responsible for the classification decision.
A survey conducted with domain experts, beginners and unskilled people proof that the usage of explanations increases the trust and confidence in the automatic decision system.
Also, an analysis of the latent space adopted by the explainer unveils that some of the most frequent skin lesion classes are distinctly separated. 
This phenomenon could derive from the intrinsic characteristics of each class and, hopefully, can provide support in the resolution of the most frequent misclassifications by human experts.
\end{abstract}


}


\section{Introduction}

Artificial Intelligence (AI) based decision support systems have recently gained a huge consideration in different domains due to their remarkable performance.
However, their adoption in sensitive scenarios that involves decision on humans, such as the medical one, has raised ethical concerns about the missing transparency for the decisions based on AI suggestions~\cite{pedreschi2019meaningful,miller2019explanation}.
The need is to develop AI systems that are able to assist doctors in taking informative decisions, complementing their own knowledge with the information and suggestion yielded by the AI system~\cite{ISIC2021,panigutti2020doctor}. 
However, if the logic for the decisions of AI systems are not available it would be impossible to accomplish this goal.
Skin image classification is a typical example of this problem.
Indeed, when effective deep learning models are adopted to solve this problem there are no clues that help in understanding the reasons for the decision outcome and that do lead to a natural interaction with the practitioner. 
Therefore, it is essential to augment currently adopted classification models with explainability components that enrich the interactions and provide the human with additional exploration and diagnosis tools \cite{aniek}.
This is the problem addressed in this paper when focusing on skin lesion diagnosis from images.

For all these reasons, eXplainable AI (XAI) has recently received much attention~\cite{adadi2018peeking,miller2019explanation,guidotti2019survey}.
Saliency maps are the type of explanation most widely returned for image classifiers.
A saliency map is an image highlighting each pixel's positive (or negative) contribution to the decision outcome.
Various approaches are proposed in the literature to explain image classifiers through a saliency maps.
First, we underline that explanation methods can be categorized as \textit{model-specific or model-agnostic}, depending on whether the explanation method exploits knowledge of the internal structure of the black-box or not; \textit{global or local}, depending on whether the explanation is provided for the black-box as a whole or for any specific instance.
Various model-specific explainers such as IntGrad~\cite{sundararajan2017axiomatic}, GradInput~\cite{shrikumar2016not}, and $\varepsilon$-LRP~\cite{bach2015pixel} are specifically designed to explain deep neural networks and return as explanation saliency maps.
The saliency maps returned by these kinds of approaches are typically scattered and not easy to read in a critical medical situation. 
On the other hand, \lime{}~\cite{ribeiro2016should} and \shap{}~\cite{lundberg2017unified} are two of the most well known model and data agnostic local explainers.
\lime{} randomly generates a local neighborhood ``around'' the instance to explain, labels them using the black-box under analysis and returns an explanation using as surrogate model a linear regressor.
\shap{} leverages game theory and exploits the Shapley values of a conditional expectation function of the black-box, providing for each feature the unique additive importance.
\lime{} and \shap{} can be applied to explain image classifiers and return explanation in the form of saliency maps.
Unfortunately, both \lime{} and \shap{} require a segmentation procedure that affects the explanation: the neighborhoods considered are no longer plausible instances but simply the image under analysis with some pixels ``obscured''~\cite{guidotti2019investigating}. 
This is also not beneficial nor trustful in a medical context.

To overcome these issue, in~\cite{guidotti2019black} has been proposed \abele{}, a local model agnostic explainer specifically designed for image classifiers.
Given an image to be explained and an image classifier, the explanation provided by \abele{} is composed of \textit{(i)} a set of \textit{exemplar} and \textit{counter-exemplar} images, and \textit{(ii)} a \textit{saliency map}.
Exemplars and counter-exemplars are images classified with the same outcome as the input, and with a different outcome, respectively, while a saliency map highlights the areas which are more responsible for the decision.  

The aim of this paper is to extend and exploit the methodology illustrated in~\cite{guidotti2019black} and~\cite{ISIC2021}, and to study the usability of an explanation method into a real medical setting.
In particular, we focus on skin lesion diagnosis from images.
We rely on the labeled dataset available from the ISIC 2019 image classification challenge.
We train on the ISIC dataset a state-of-the-art deep learning classifier using the ResNet CNN architecture, i.e., a ResNet~\cite{he2016resnet}.
After that, we explain the classifier decisions through \abele{}~\cite{guidotti2019black}.
In this way, the practitioner can easily reason on top of the exemplars and counter-exemplars returned by the explainer.
Our goal is to assess to which extent these exemplar and counter-exemplar explanations are effectively useful through a user study involving humans.
We design the experiment as a survey where participants are asked to address certain tasks on the basis of the classification outcome and of the explanations. 

The survey has been conducted with domain experts, beginners and unskilled people.
The results of the survey show that the usage of explanations increases the trust and confidence in the automatic decision system. 
This phenomenon is more evident among domain experts and people with the highest level of education. 
Interestingly, the older segment of the population shows a chronic mistrust of AI models that is unaffected by the exposition of \abele{} explanations.
Also, we observe that after receiving wrong advice by an AI model, domain experts tend to decrease their trust in the same model for future analysis.
As additional result we highlight the analysis of the latent space of the autoencoder made available by \abele{}.
The latent space analysis suggests a interesting separation of the images that can hopefully be helpful in separating apart similar classes of skin lesion that are frequently misclassified by humans.
Finally, we also design and develop a user interaction module to analyze the explanations returned by \abele{}.

The rest of the paper is organized as follows.
Section~\ref{sec:method} presents the methodology adopted.
In Section~\ref{sec:casestudy} we illustrate in detail the settings of the case study addressed, and we present the interaction module.
Section~\ref{sec:results} illustrates the survey and shows the results obtained, while Section~\ref{sec:latent} presents the analysis of the latent space.
Finally, Section~\ref{sec:conclusion} summarizes the contribution and proposes future research directions.

\section{Methodology}
\label{sec:method}
In this section we briefly present the two main components of the methodology adopted to classify and explain the ISIC dataset.
Details can be found in~\cite{he2016resnet,guidotti2019black,ISIC2021}. 

\subsection{ResNet Classifier}
In order to provide a classifier that performs sufficiently well for downstream learning steps, we choose to train a Neural Network (NN) with an architecture powerful enough to accomplish image classification. 
In particular, we selected a \textit{ResNet}, a NN architecture providing validated performance on many complex datasets and tasks~\cite{he2016resnet}.
Instead of training the ResNet from scratch, we choose to perform a \textit{transfer learning} task with a ResNet pretrained on the ImageNet dataset. 
This training strategy is largely applied for the case where the number of data is limited with respect to the complexity of a NN~\cite{pan2010transfer}. 
To perform the transfer learning, we replace the last fully connected layer with the newly initialized one. 
The number of output dimension is adapted to the number of classes in the dataset. 
Then, the classification layer is learned from scratch and the rest of the ResNet is fine-tuned. 
As loss function we adopt a binary cross entropy loss for each class, so that the task can be considered as individual one-vs-rest binary classification problem.

\begin{figure}[t]
    \centering
    \includegraphics[width=\columnwidth]{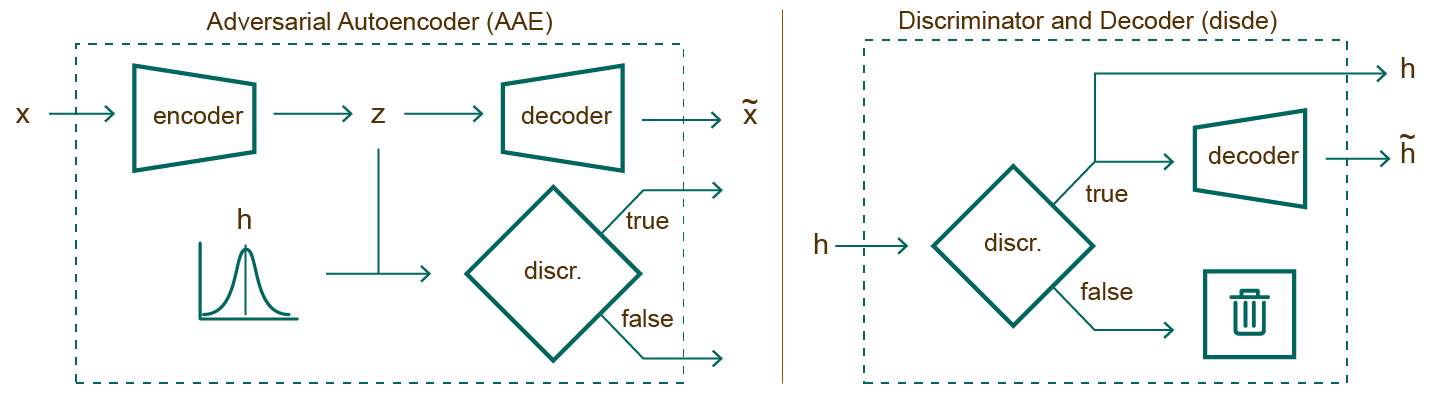}
    \caption{\textit{Left}: AAE architecture. \textit{Right}: Discriminator and Decoder module.}
    \label{fig:AAE}
\end{figure}

\begin{figure}[t]
    \centering
    \includegraphics[width=\columnwidth]{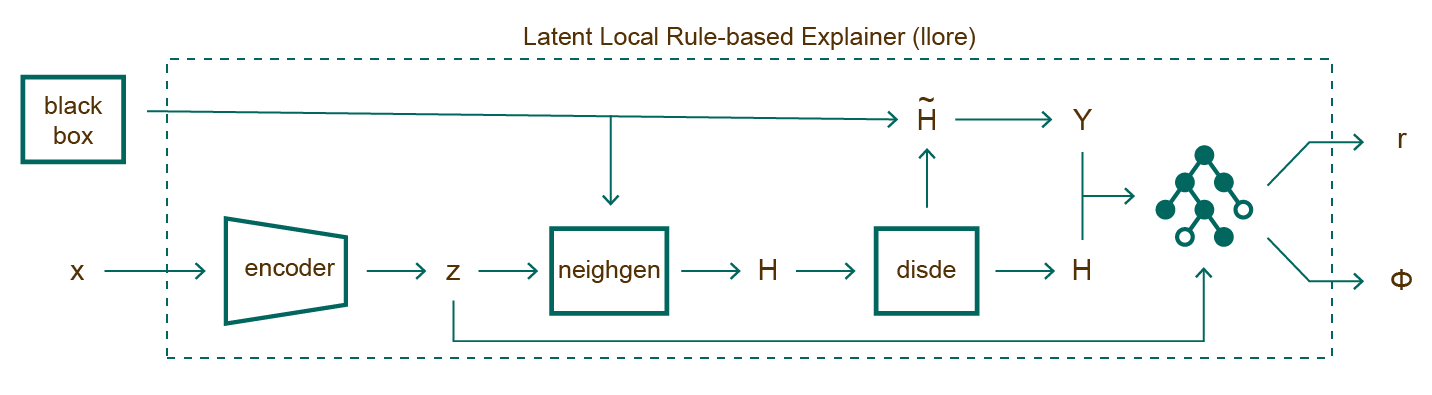}
    \caption{Latent Local Rules Extractor ($\mathit{llore}$) module.}
    \label{fig:learning}
\end{figure}

\subsection{ABELE Explainer}
As explainer we adopted \abele{} (Adversarial black-box Explainer generating Latent Exemplars)~\cite{guidotti2019black}.
\abele{} is a local model agnostic explainer that takes as input an image $x$ and a black-box classifier $b$, and returns \textit{(i)} a set of \textit{exemplar} and \textit{counter-exemplar} images, and \textit{(ii)} a \textit{saliency map}.
Exemplars and counter-exemplars are images synthetically generated and classified with the same outcome as $x$, and with an outcome other than $x$, respectively. 
They can be visually analyzed to understand the reasons for the decision. 
The saliency map highlights the areas of $x$ that contribute to its classification and areas that push it into another class.

In short, \abele{} works as follows.
First, it generates a neighborhood $H$ in the latent feature space exploiting an AAE~\cite{makhzani2015adversarial}.
The AAE architecture (Fig.~\ref{fig:AAE}-left) includes an \texttt{encoder}: $\mathbb{R}^n \rightarrow \mathbb{R}^k$, a \texttt{decoder}: $\mathbb{R}^k \rightarrow \mathbb{R}^n$ and a \texttt{discriminator}: $\mathbb{R}^k \rightarrow [0, 1]$ where $n$ is the number of pixels in an image and $k$ is the number of latent features.
The image $x\in\mathbb{R}^n$ to be explained is passed as input to the AAE where the \texttt{encoder} returns the latent representation $z \in \mathbb{R}^k$ using $k$ latent features with ${k \ll n}$. 
The neighborhood generation of $H$ (\texttt{neighgen} module in Fig.~\ref{fig:learning}) may be accomplished using different strategies.
In our experiments, we adopt a genetic approach maximizing a fitness function like in~\cite{guidotti2019factual}. 
The module in Fig.~\ref{fig:learning} is named \textsc{llore}, as a variant of \lore{}~\cite{guidotti2019factual}.
After the generation process, for any instance $h \in H$, \abele{} exploits the \texttt{disde} module (Fig.~\ref{fig:AAE}-right) for both checking the validity of $h$ by querying the \texttt{discriminator} and decoding it into $\widetilde{h}$.
Then, it queries the black-box $b$ with $\widetilde{h}$ to get the class $y$, i.e.~$b(\widetilde{h})=y$.
Given the local neighborhood $H$, \abele{} builds a decision tree classifier $c$ trained on $H$ labeled with $b(\widetilde{H})$. 
The surrogate tree is intended to locally mimic the behavior of $b$ in the neighborhood $H$.
It extracts the decision rule $r$ and counter-factual rules $\Phi$ enabling the generation of \textit{exemplars} and \textit{counter-exemplars}.
Fig.~\ref{fig:learning} shows the process that, starting from the image to be explained, leads to the decision tree learning, and to the extraction of the decision and counter-factual rules. 

The overall effectiveness of \abele{} lies in the goodness of the encoder and decoder function adopted.
In other words, the better is the AAE, the more realistic and useful will be the explanations.
In the next section we highlight some peculiarities of the structure of the AAE required to obtain reliable results for the ISIC dataset.

\begin{figure}[t]
    \centering
    \includegraphics[width=\columnwidth]{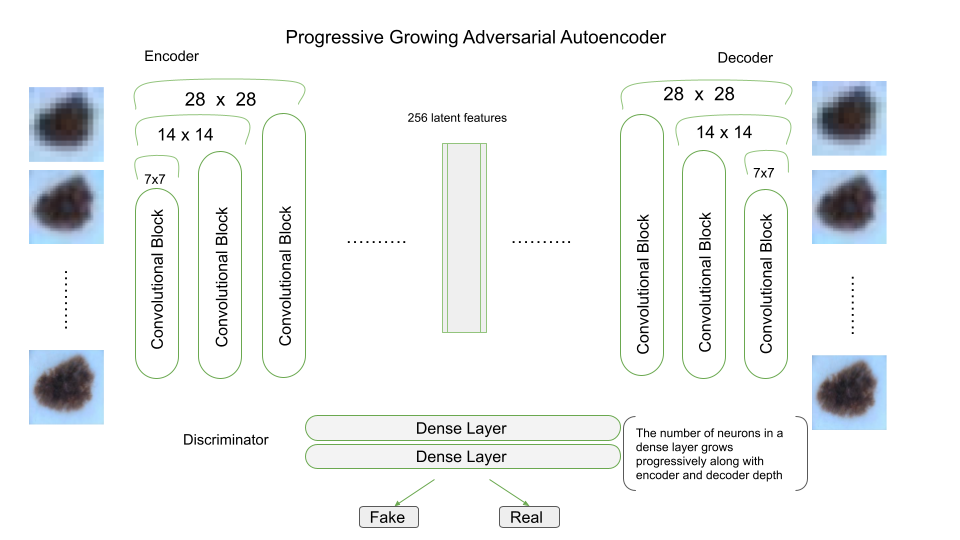}
    \caption{A Progressive Growing AAE.}
    \label{fig:PGAAE}
\end{figure}

\subsection{Progressive Growing AAE}
We summarize here the customization of \abele{} we carried on in order to make it usable for the complex image classification task of the ISIC dataset.
Details can be found in~\cite{ISIC2021}.
Generative Adversarial models are generally not easy to train as they are usually affected by a number of common failures. 
These problems vary from a diversified spectrum of \textit{failures in convergence} to the famous \textit{mode collapse}~\cite{modecollapse}, the tendency by the generator network to produce a small variety of output types.
Such problems mainly arise from the competing scheme generator and discriminator are trained on. 
In addition, we often face the further complication to deal with real world datasets that are far from ideal: fragmentation, imbalance, lack of uniform digitization, shortage of data are primary challenges of big data analytic for healthcare. 
Training an AAE in a standard fashion to reproduce samples from ISIC dataset without taking special care of all issues mentioned above resulted in extremely poor performance, mostly due to a persistent collapse mode. 

In order to overcome such generative failure and dataset limitations, we implemented a collection of cutting edge techniques that altogether are capable of addressing all the issues we mentioned and successfully training an AAE with adequate performance.
In particular, we address mode collapse using ad-hoc tricks like Mini Batch Discrimination~\cite{minibatch} and Denoising autoencoders~\cite{denoising}.
As model of AAE, we adopted a Progressive Growing AAE.
Progressive Growing GANs~\cite{progressive} have been introduced as an extension of GANs. 
Progressive Growing helps to achieve a more stable training of generative models for high resolution images like in our case. 
The main idea is to start with a very low resolution image and step by step adding block of layers that simultaneously increase the output size of the generator model and the input size of the discriminator model until the desired size is achieved.
However, while in a GAN, the discriminator is linked to the generator output, in an AAE, the discriminator takes as input the encoded latent space instead of the full reconstructed image. 
Thus, in~\cite{ISIC2021} we define Progressive Growing Adversarial Autoencoder (PGAAE) as follows.
Starting with a single block of convolutional layers for encoder and decoder, we are able to reconstruct low resolution images (7$\times$7 pixels), then step by step we increase the number of blocks until the networks are powerful enough to manage images of the desired size, i.e., 224$\times$224 pixels in our case. 
The latent space dimension is kept fixed, consequently the discriminator takes as input tensors always of the same size. 
Although one could fix also the network of the discriminator, we found helpful to progressively increasing also the width of this network so that the discriminator can deal each step with a more structured information. 
The incremental addition of the layers allows the PGAAE to first learn large scale structure and progressively shift the attention to finer detail. 

The PGAAE network is reported in Figure~\ref{fig:PGAAE}.
In our implementation we used six blocks of layers in order to have a trained AAE able to reproduce skin lesion images of size 224x224 pixels.
In summary, as learning tricks we relayed on minibatch discrimination, denoising and progressive structure.
Thanks to the PGAAE, mode collapse is greatly reduced, and we are able to generate variegate and good quality synthetic skin lesion images with \abele{} acting as exemplars and counter-exemplars.

\section{Case Study: Skin Lesion Diagnosis}
\label{sec:casestudy}
In this section we illustrate the case study analyzed presenting details for the dataset, and for the training of the classifier, and of the autoencoder.

\subsection{Dataset}
The International Skin Imaging Collaboration (ISIC), sponsored by the International Society for Digital Imaging of the Skin (ISDIS) proposed
the \textit{skin lesion analysis towards melanoma detection challenge} to improve the international effort in melanoma diagnosis \footnote{https://challenge2019.isic-archive.com/}.
The challenge consists in  developing a classifier to recognize among nine different diagnostic categories of skin cancer: MEL (Melanoma), NV (Melanocytic nevus), BCC (Basal cell carcinoma), AK (Actinic keratosis), BKL (Benign keratosis), DF (Dermatofibroma), VASC (Vascular lesion), SCC (Squamous cell carcinoma), UNK (Unknown, none of the others / out-of-distribution).
The dataset is composed of a training set of 25,331 images of skin lesions and their category (labels); a test set of 8,238 images of which the label is not publicly accessible.

\subsection{ResNet Training}
We separated the training into two parts: 80\% samples used for training and 20\% for validation.
Since the UNK category is a reject option and is not available in the training we focused on the 8 other categories.
Also, since images have different resolutions, we applied the following preprocessing:
\begin{itemize}
    \item For the training, the images are randomly rescaled, rotated and cropped to generate the input to the network\footnote{Such preprocessing does not deform the lesions in the image.}.  Resolution of the preprocessed images is 224$\times$224.
    \item For the validation and test, each image is firstly rescaled to 256$\times$256 according to the shorter edge, then cropped at the center into a 224$\times$224 image.
\end{itemize}

For the evaluation, we used the same metric adopted in the submission system of ISIC 2019.
The model is evaluated on (our) test set with the normalized (or balanced) multi-class accuracy defined as the average of recall obtained in each class. 
This metric makes all the classes equally important to avoid that classifier performs well only for dominant classes.
The trained ResNet model achieves 0.838 of balanced multi-class accuracy on the test set.
%


\subsection{PGAAE Training}
A customization of \abele{} was necessary in order to make it usable for the complex image classification task addressed by the ResNet classifier. 
Details are available in~\cite{ISIC2021}. 
After a thorough fine tuning of all three networks structures (encoder, decoder and discriminator) our PGAAE with $256$ latent features achieves a reconstruction error measure through RMSE that ranges from $0.08$ to $0.24$ depending on whether we consider the most common or the most rare skin lesion class.
We selected 256 as number of latent features because, from preliminary experiments, it was the number that simultaneously guaranteed a good reconstruction error, a good resolution of the images and did not involve an excessive waste of computational resources.
Also, for images of the desired size, i.e., 224$\times$224, it is common in literature to choose a number of latent features that varies between 64 and 512.
Data augmentation was necessary to overcome scarcity and imbalance of the dataset. Mode collapse was greatly reduced, and we were able to generate variegate and good quality skin lesion images. 
\abele{} equipped with such PGAAE is able to generate meaningful explanations and it can be tested in a survey involving real participants described in the next section.

\begin{figure}[t]
    \centering
    \includegraphics[width=\columnwidth]{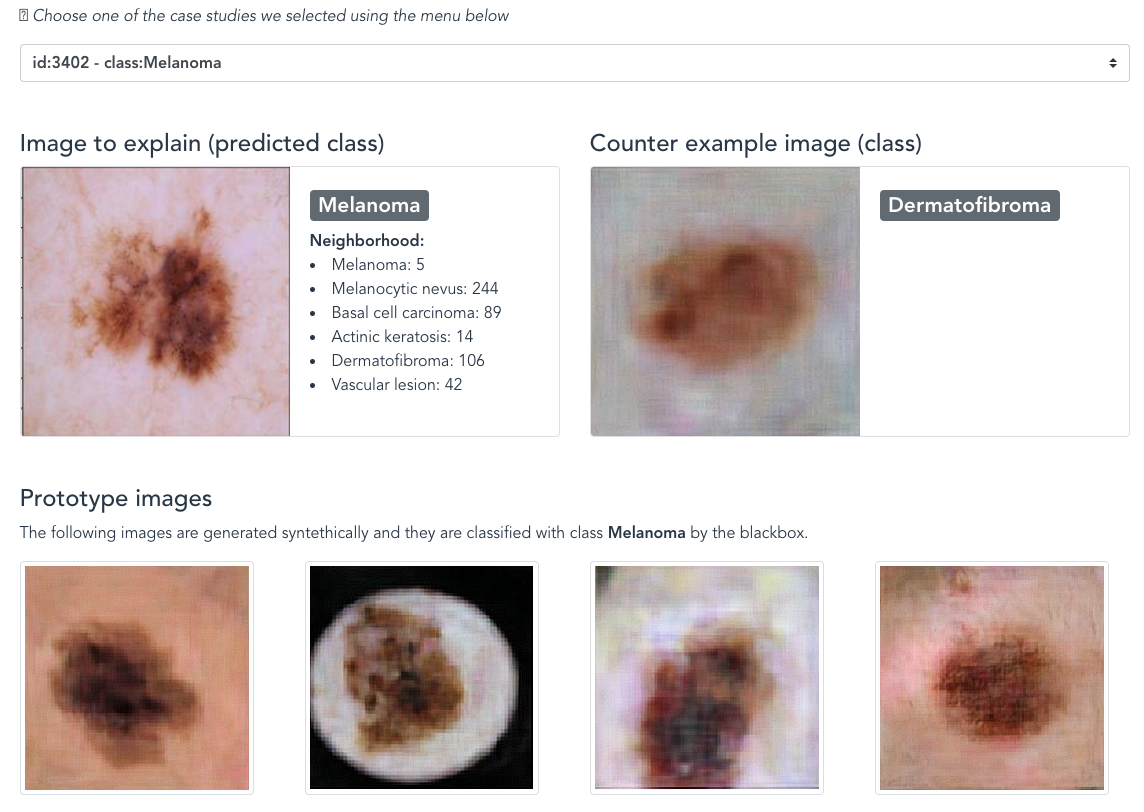}
    \caption{User interaction module to present the results of classification and the corresponding explanation. The upper part presents the input instance and a counter-exemplar. The lower part shows for exemplars that share the same class as the input.}
    \label{fig:isic_viz}
\end{figure}

\subsection{ABELE User Interaction Module}
We present here the novel interaction module for the explanations returned by \abele{}. 
The module facilitates the comprehension of both the suggestion of the black-box model and the explanation produced by \abele{}.
Figure~\ref{fig:isic_viz} shows a screenshot from a web application\footnote{\url{https://kdd.isti.cnr.it/isic\_viz/}} we implemented to present to the user the outcome of our system on a specific instance.
The visual space is organized in two sections. 
The upper part shows the instance under analysis $x$ with the classification returned by the ResNet on the left, and a synthetic counter-exemplar image returned by \abele{} on the right. 
In Figure~\ref{fig:isic_viz} we have a Melanoma under analysis and a quite similar synthetic image classified as Dermatofibroma.
Still on the left is reported a list of instances that are similar to the instance under analysis but with a different label.
This list, that is the neighborhood generated by \abele{} through the PGAAE, gives an indication of the variety of the instances in the latent space around the image analyzed. 
The counter-exemplar is selected among this list as the image that minimizes a distance similarity with $x$ and maximizes the classification similarity but w.r.t. a different label.
Finally, the lower part of the module shows four exemplars, i.e. a set of images returned by \abele{} that have the same label assigned by the ResNet to $x$.

The ABELE interaction module is implemented in Javascript as a web application.
More in detail, it communicates with a backend that exposes the functionalities of the ResNet and of \abele{} by means of a RESTful interface. 
We implemented and deployed a demonstator of the system by letting the user choose from a set of instances, instead of uploading a new one. 
This demostrator has been used for the preparation of the survey presented in the next section.

\section{Validation and Survey}
\label{sec:results}
We designed a survey with the idea of validating the effectiveness of the explanations returned by \abele{} for skin lesion diagnosis.
In particular, the main purpose is to validate the effectiveness of the explanations in assisting doctors and medical experts in the diagnosis and treatment of skin cancers, as well as to investigate their confidence in automatic diagnosis models based on black-boxes and on the explanations provided by the explainer.

\subsection{Survey Structure}
The survey is organized in ten questions composed of various points with the same general structure described in the following.
We highlight that, if not specified, for every point a different image $x$ is provided to the user.
In the remaining of the treatment we name each question Q$i$ with $i \in [1,10]$.

\textit{Point 1 (P1).} 
The participants are presented with an unlabeled skin lesion image randomly chosen among the dataset and its explanation as generated by \abele{} and presented by the interaction module. 
In particular we presented to the user two exemplars and two counter exemplars of another lesion class.
Participants were asked to classify the given image among two different given classes exploiting the explanation.
The idea behind this point is to understand if the explanations returned by \abele{} really helps in separating between different images, even for non-expert users.
From another perspective, this can be considered as the human evaluation of the \textit{usefulness} metric synthetically observed in~\cite{guidotti2019black}.

\textit{Point 2 (P2).}
The participants are presented with a labeled image and they are asked to quantify the level of confidence in the black-box classification (using a 0–100
slider). 

\textit{Point 3 (P3).} 
The participants are presented with the same labeled image of P2, but this time with the visual aid of the explanation returned by \abele{}, and they are asked to quantify his confidence once more after looking at the explanations.
The objective of this couple of points is to understand if there is an increase/decrease in the confidence towards the AI after having observed an explanation.

\textit{Point 4 (P4).}
The participants are asked to quantify how much the exemplars and counter-exemplars helped them to classify skin lesion images in accordance with the AI, and how much they trust the explanations produced by \abele{}.

During the survey, participants are not informed of the correctness of their prediction nor they received further suggestions by looking back at their previous answers or explanations.
In order to investigate recipient's reaction to an incorrect advice, in Q6 we intentionally entered a wrong classification (P2) followed by further wrong advice concerning exemplars and counter-exemplars (P3). 
All the other nine instances were correctly presented with the ground truth label.

\subsection{Hypothesis and Objectives}
The structure of the investigation reflects the following hypothesis we intend to address.

\begin{itemize}
    \item \textit{H1:} The explanations returned by \abele{} help the recipients in the classifications task, especially domain experts which are supposed to achieve a higher classification score (implicit assessment through P1).
    \item \textit{H2:} The explanations returned by \abele{} improve the recipients trust and confidence toward the black-box classification (implicit assessment through P2, and P3, explicit through P4).
    \item \textit{H3:} After receiving a wrong advice from an AI, participants will show a substantial decline in confidence and trust toward that model (implicit assessment through the error inserted).
\end{itemize}

\begin{figure}[t]
    \centering
    \includegraphics[width=\columnwidth]{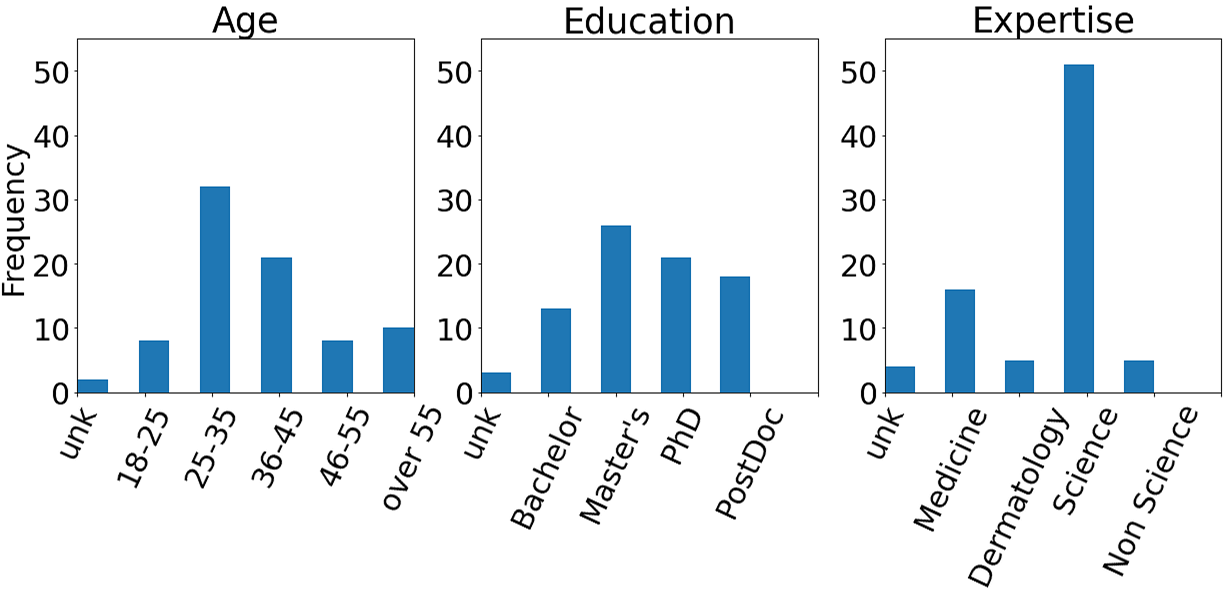}
    \caption{Aggregate demographic statistics of the survey participants.}
    \label{fig:anagraphic}
\end{figure}

\subsection{Survey Results}
A total of 156 participants completed the study. 
Participants signed up for the survey online, after digitally signed a consent form followed by a short demographic survey and a brief introduction about all the different types of skin lesion cancer involved in the process.
Since participants were not forced to answer to all the ten questions, we chose to consider only who had completed at least 27\% of the questionnaire, corresponding to at least an entire question with an answer for each point. 
Aggregate demographic statistics of the participants are available in Figure~\ref{fig:anagraphic}. 
The vast majority of participants have a scientific background, among which a fair number having completed studies in medicine or dermatology. 

First, we have attached to each participant a score that measures their performance in P1 that test their ability to classify skin lesion images exploiting the explanation.
We divided participants into two sub-samples.
Sub-sample A contains participants who achieved a score of at least 70\% images correctly classified, and sub-sample B contains participants who achieved a score lower than 70\%.
Different threshold in the range $[60\%-90\%]$ were also considered but discarded since they yielded approximately the same statistics results. 
Although the average performance in itself is remarkable (score 82.02\%) even among people who were not domain experts nor had specific medical knowledge (score 78.67\%), we are interested in the sub-sample of people specialized in medicine or more specifically in dermatology (score 91.26\%). 
A one-way ANOVA on ranks (Kruskal–Wallis H test)~\cite{ANOVA} applied to sub-sample A and B shows no significant effects on classification performance given by education level nor age distribution.
However it shows a pronounced effect ($F = 4.061$, $p = 0.043$) given by the specific field in which the participants specialize - participants with a medicine degree and/or a dermatology specialization are more frequent among people in sub-sample A. 
Thus, H1 is supported at least for domain experts - however also participants from other domains showed a remarkable performance.

\begin{figure}[t]
    \centering
    \includegraphics[width=\columnwidth]{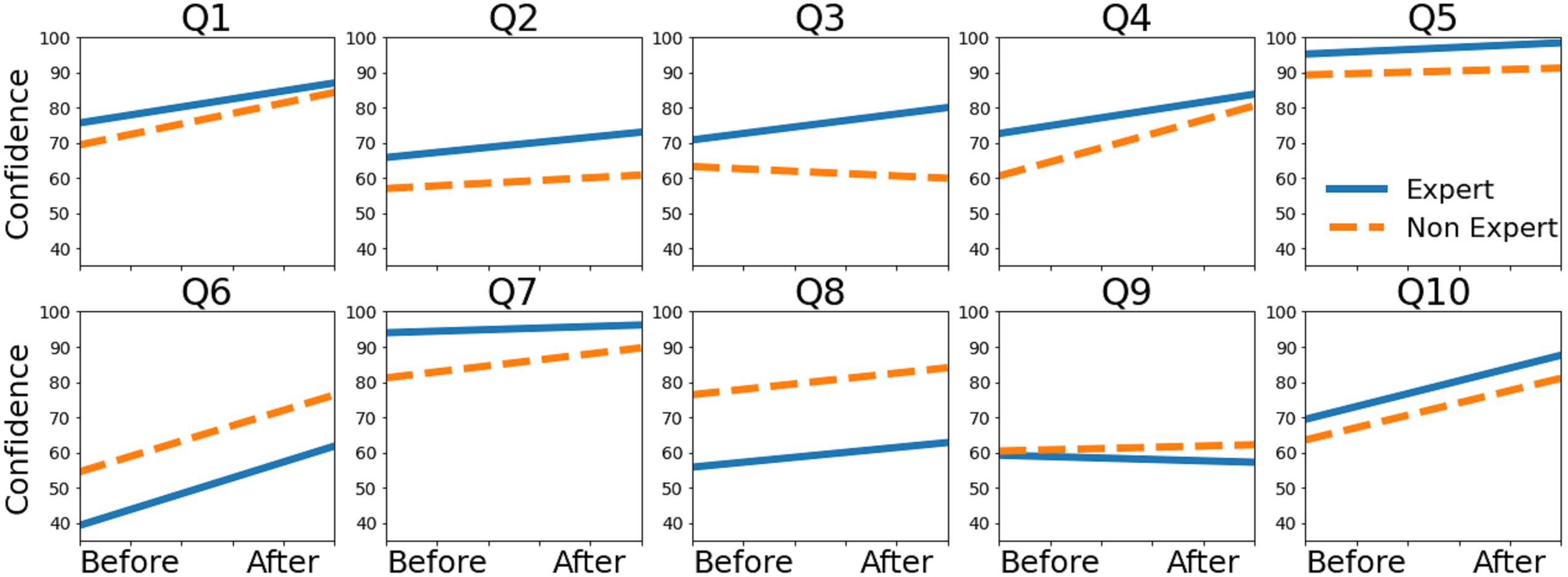}
    \caption{Participants confidence in the classification of the black-box before and after receiving the explanation of \abele{}.}
    \label{fig:trust}
\end{figure}

Figure~\ref{fig:trust} shows the participants' confidence in the black-box classification before and after looking at the explanations returned by \abele{}, i.e., reports the responses obtained for P2 and P3. 
Except for Q3, all the other questions show a significant increase of trust after looking at exemplars and counter-exemplars, i.e., an increase between P2 and P3, indicating that, in general, the explanations helps in increasing the model trust. However, Figure \ref{fig:trust} suggests that Q3 anomaly was caused by the sub sample of non medical expert.
Such an increase of confidence from 67.69\% to 77.12\% , is maximum for Q6 (+21.95\%), the only one which was misclassified by the black-box.
Moreover, Q6 presents a confidence, prior to the explanations, lower than all the others, i.e., of 53.08\%. 
A plausible reason for this phenomenon could be a substantial decline in participant's confidence following a wrong suggestion, which is then completely recovered if the first wrong suggestion is followed by other consistent wrong suggestions.
Participants reject a wrong advice but tend to adapt and reset their knowledge baseline if the wrong suggestions consistently repeat themselves.

%
%

\begin{figure}[t]
    \centering
    \includegraphics[width=\columnwidth]{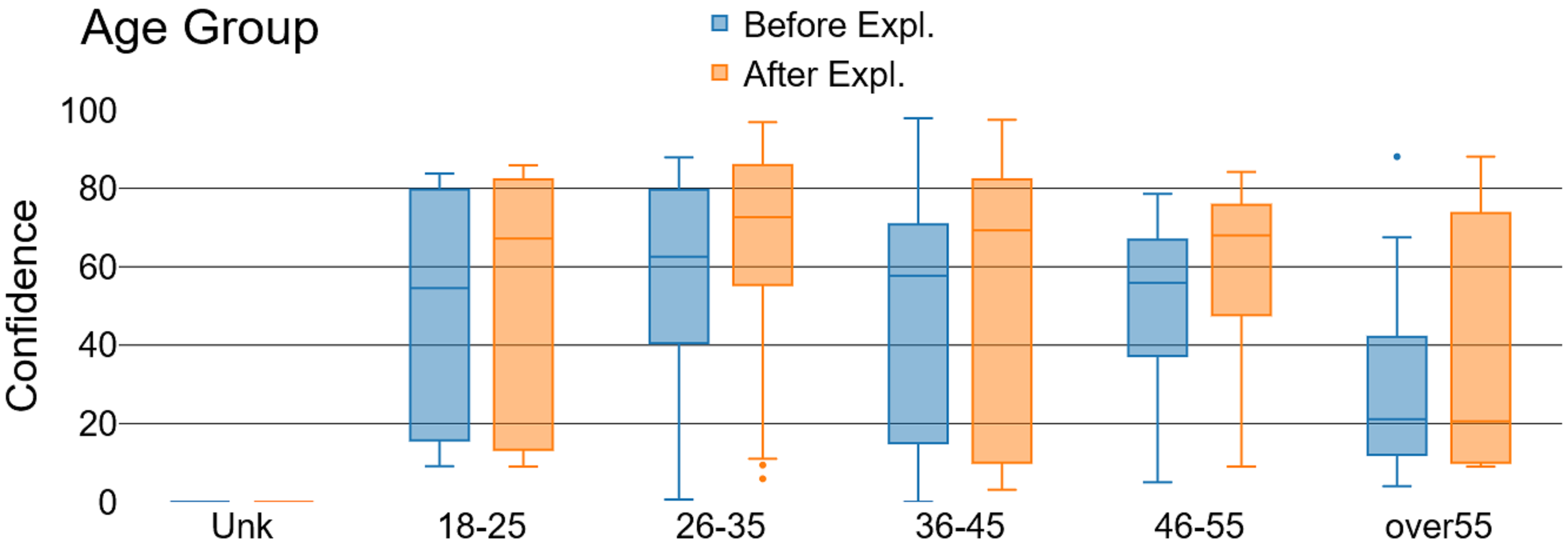}\hspace{2mm}\\
    \includegraphics[width=\columnwidth]{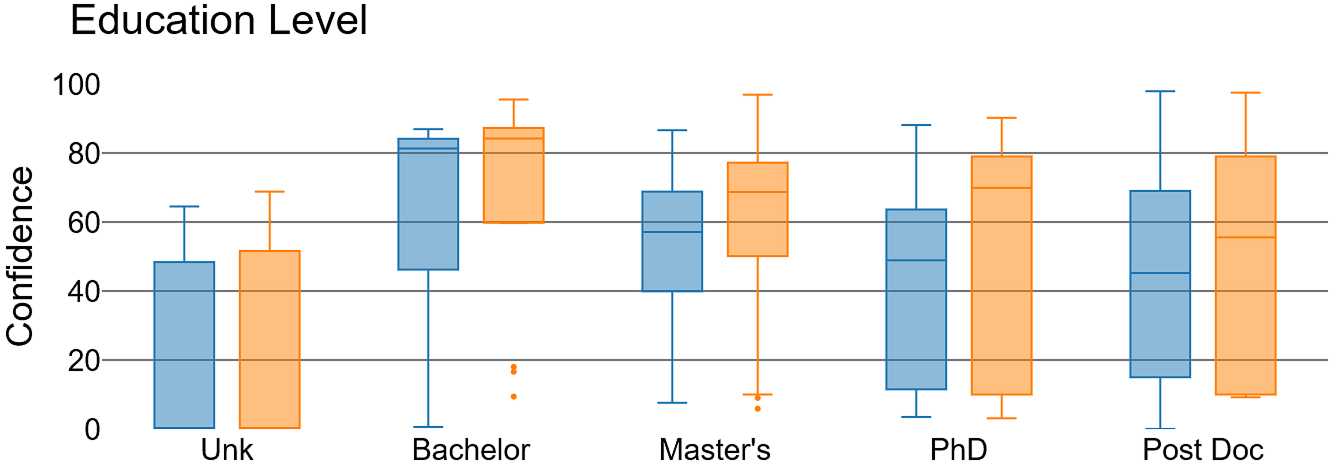}\hspace{2mm}\\
    \includegraphics[width=\columnwidth]{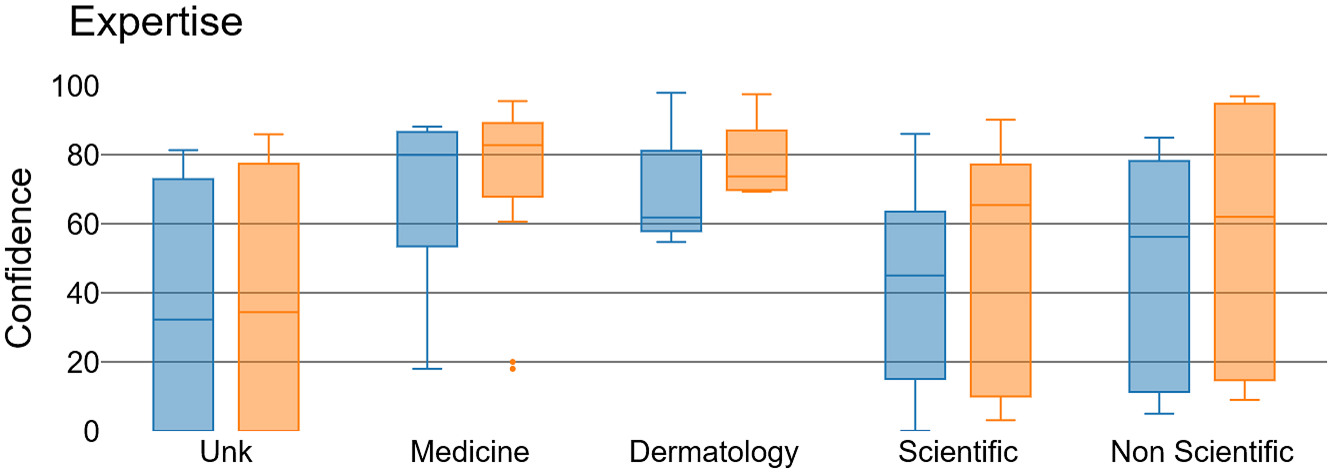}
    \caption{Participants' confidence among different age groups (top), education level (center), domains (bottom), before and after explanations.}
    \label{fig:trust_boxplot}
\end{figure}

Increase in confidence was not uniform among all participants. 
The results summarized in Figure~\ref{fig:trust_boxplot} seem to suggest the following aspects. 
First, there is a not negligible increase in confidence among all ages except for age group over 55, for which not only the confidence is very low in itself but even decreases after having benefited from the explanations.
This may be caused by the fact that the older segment of the population has an inherent distrust of AI models in general, while younger sections of the population are mentally more open to such models.
Second, the confidence before looking at the explanations decreases as the level of study increases, while more educated participants show a notable increase after the explanations (a possible reminiscence of the Dunning-Kruger effect~\cite{dunning}).
Third, as expected, the confidence level is much higher for people belonging to the medical domain than for participants from other scientific disciplines and even more so for those specializing in non-scientific disciplines.

\begin{figure}[t]
    \centering
    \includegraphics[width=\columnwidth]{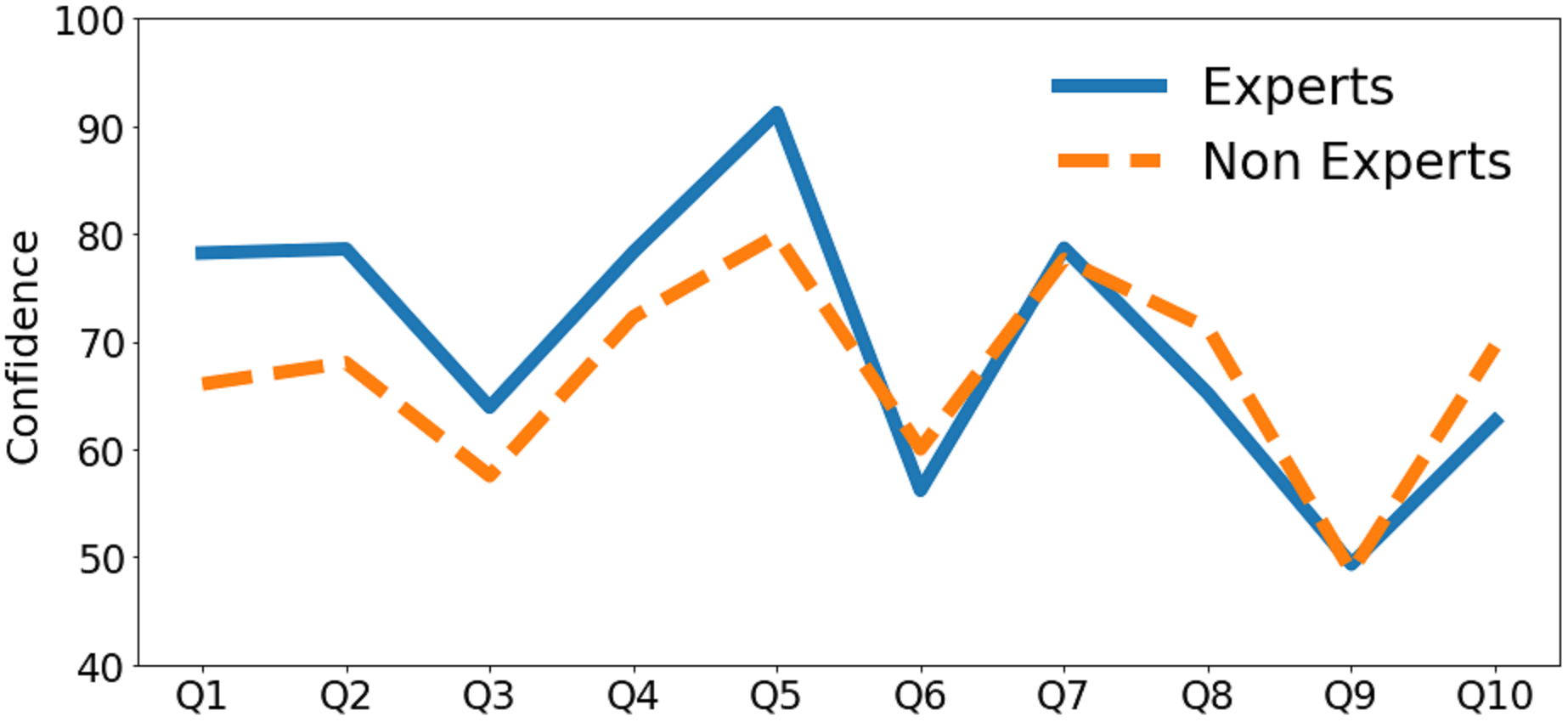}
    \caption{Participants confidence towards \abele{} explanations.}
    \label{fig:trustQ6}
\end{figure}

As mentioned earlier, Q6 was specifically chosen from those misclassified by the black-box, in order to investigate participants' reaction and behavior in that and subsequent instances (H3).
The results show a slight mistrust toward the sixth black-box classification, although there is no statistically significant drop in confidence after receiving wrong advice by an AI model (68.75\% for Q1 to Q5, 60.03\% for Q6 and 66.71\% for Q7 to Q10). 
On the contrary, if we restrict our study to the sub-sample of medical experts, Figure~\ref{fig:trustQ6} shows a 14\% drop of confidence after receiving the wrong advice (78.04\% for Q1 to Q5, 56.19\% for Q6 and 63.95\% for Q7 to Q10), supporting H3: after receiving wrong advice from an AI model, domain experts show a decline in confidence and trust toward that model in subsequent instances.

\begin{figure}[t]
    \centering
    \includegraphics[width=\columnwidth]{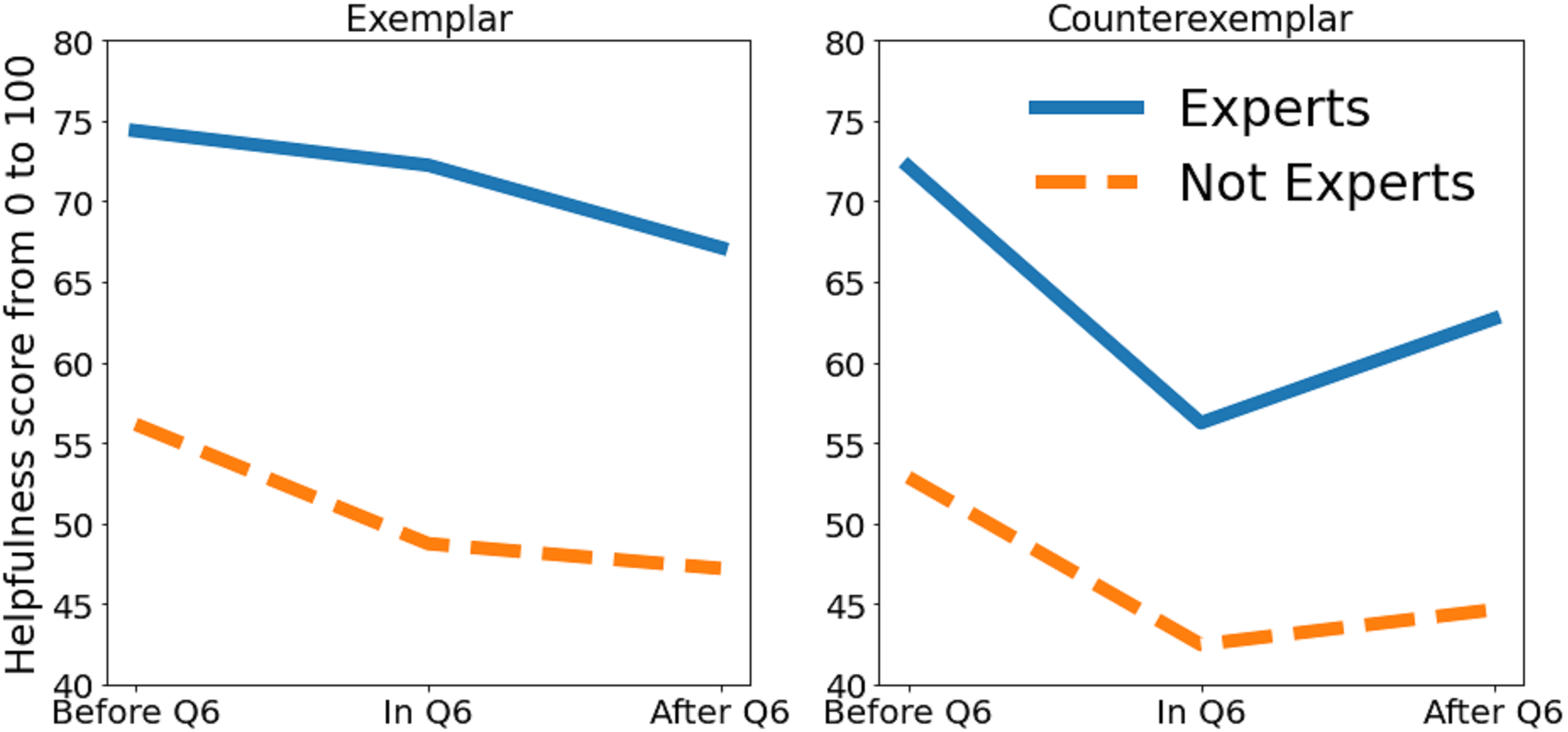}
    \caption{How much exemplars and counter-exemplars helped according to the participants' responses, divided between groups of experts and non-experts.}
    \label{fig:ex_adn_cex}
\end{figure}

Finally, in Figure~\ref{fig:ex_adn_cex} we summarize how exemplars and counter-exemplars helped participants in recognizing lesion classes, as stated by respondents in P4. 
We note here the same trend observed in analyzing confidence in explanations before and after Q6: both experts and non experts show a notable decline in confidence about exemplars and counter-exemplars helping them in the classification task. 
As expected, the explanations returned by \abele{} are considerably more helpful for medical experts than the general population. 
Moreover, we observe on the sidelines how exemplars have been more effective than counter-exemplars for both experts and for the general population. 
It can be argued that such behavior comes from having considered an 8-class classification task - in a binary classification task, exemplars and counter-exemplars may encompass a similar significance, while, as the number of classes increases, the relative importance of examples tends to grow.

\begin{figure}[t]
    \centering
    \includegraphics[width=\columnwidth]{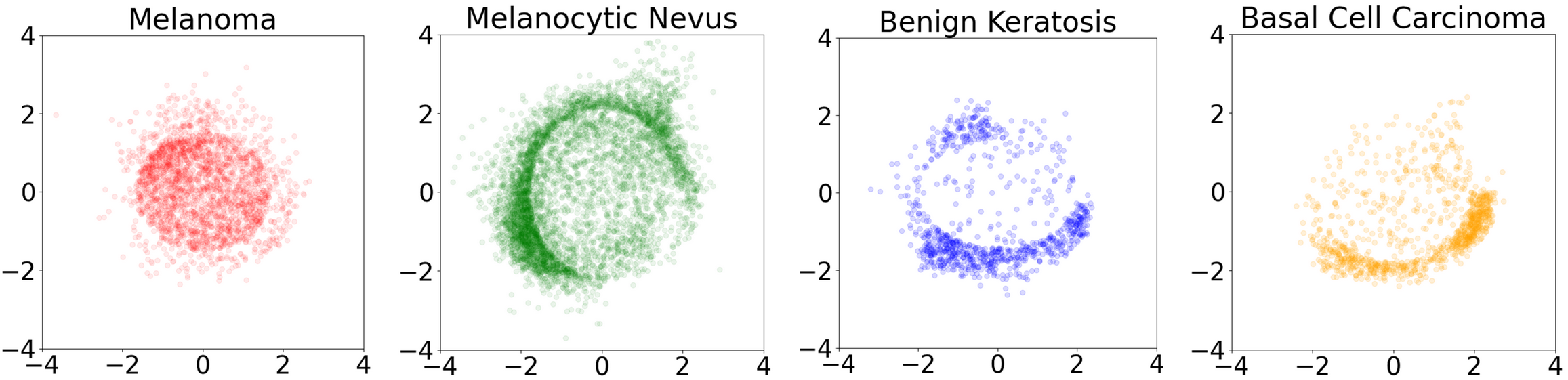}
    \includegraphics[width=\columnwidth]{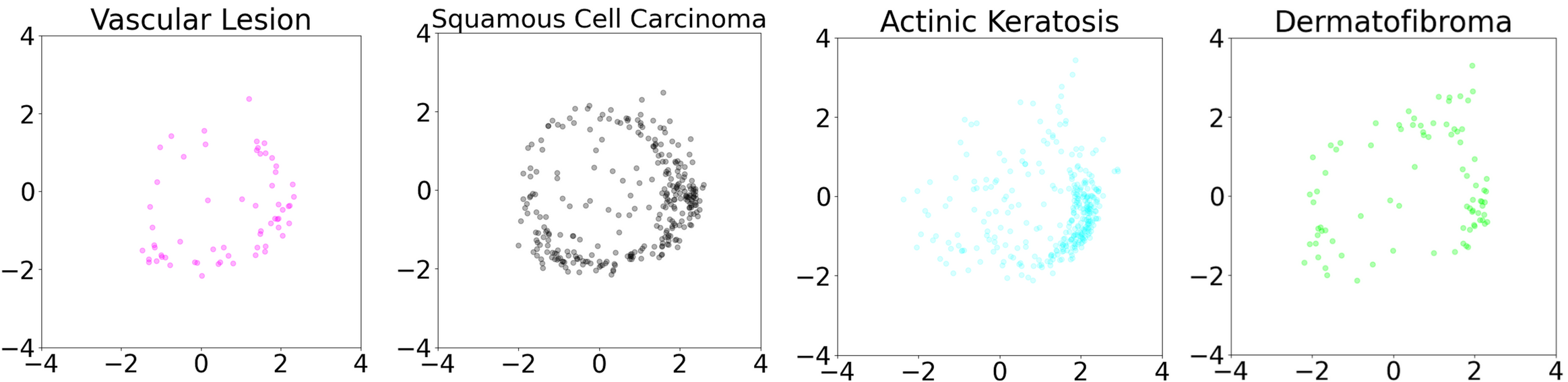}
    \caption{Skin images in the training set represented in two dimensions through a MDS applied on the latent space learned by the PGAAE.}
    \label{fig:latent1}
\end{figure}

\section{Explaining through Latent Space Analysis}
\label{sec:latent}
The PGAAE trained for \abele{} projects the ISIC dataset into a 256-dimensional latent space with a coherent posterior distribution. 
Indeed, as shown in~\cite{guidotti2019black}, as a side effect of \abele{}, we can exploit the latent space to visualize the level of proximity of individual instance of the dataset and gain useful insights.
In particular, we believe that such visual aid can help medical expert and data scientist to better understand different skin cancer characteristics and exploit it to further improve the classification performance, or the trust in the explainer. 

We adopt a Multidimensional Scaling (MDS)~\cite{MDS} as a form of non-linear dimensionality reduction to translate information about pairwise distances among latent projections into a configuration of the same cardinality mapped into a 2D Cartesian space.
Thus, through MDS we turn the latent space with 256 dimension into a visual space with 2 dimensions.
Figure~\ref{fig:latent1} shows the latent encoding of 8 skin cancer classes. 
It is worth noting that some primary features of benign and malignant cancer can be also retrieved from such 2D projection. 
Indeed, from Figure~\ref{fig:latent1} it is clear that all skin lesion classes expect \textit{Melanoma} tend to avoid the center of each diagram and accumulate over a circle, while \textit{Melanomas}, the most dangerous skin cancer, reside in the center of the plot. 

\begin{figure}[t]
    \centering
    \includegraphics[width=\columnwidth]{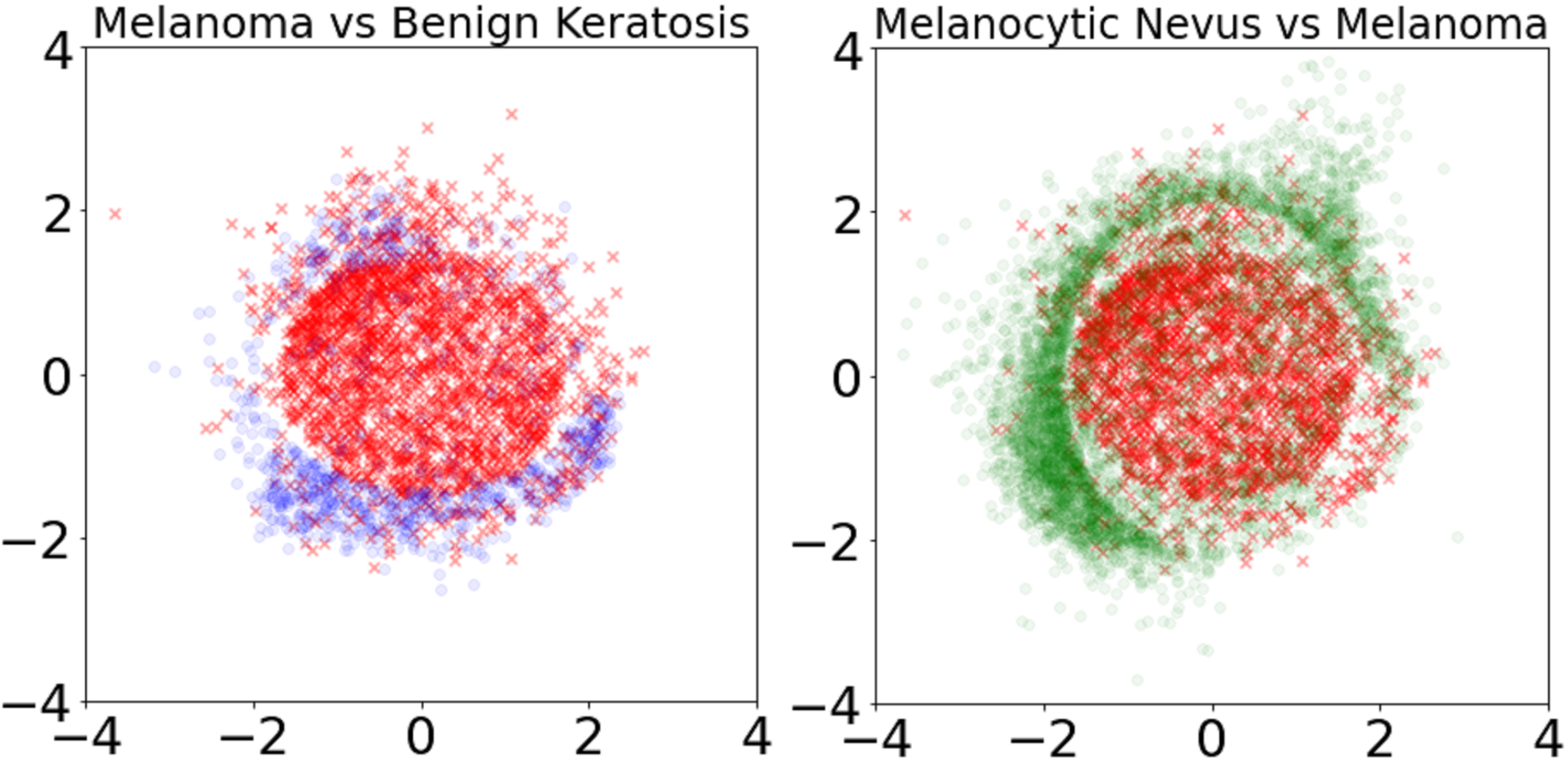}
    \caption{Visual separation between Melanoma and Benign Keratosis (Left) and Melanocytic Nevus (Right).}
    \label{fig:separation}
\end{figure}

It can be argued that such behavior is related to the similarity between these skin cancer classes. 
In~\cite{keratosis_melanoma} authors state that \textit{Benign Keratosis} is one of the lesions for which melanoma is commonly misdiagnosed, this error occurred in 7.7\% to 31.0\% of cases, depending on the study. 
Thus, we decided to train a Random Forest (RF) classifier~\cite{ho1995random} with 500 estimator trees over the 2D MDS space.
The RF classifier is able to separate apart \textit{Melanoma} from \textit{Benign Keratosis} with $85,60$\% accuracy (see Figure \ref{fig:separation}\textit{-left}).
Another important class is indeed \textit{Melanocytic Nevus} that shows a peculiar features that can be partially justified by the odd representation in Figure~\ref{fig:latent1} (2nd plot).
Here many samples still reside at the center of the plot: from 30\% to 50\% of all melanomas and more than half of those in young patients evolve from initially benign nevi~\cite{melanoma_nevus}.
The RF classifier trained over the 2D MDS space is also able to separate \textit{Melanoma} from \textit{Melanocytic Nevus} with $78,53$\% accuracy (see Figure~\ref{fig:separation}\textit{-right}).
These performances are comparable with the original black box accuracy for the overall scores in the ISIC 2019 challenge~\cite{ISIC2021}, and state of the art classification accuracy with deep convolutional neural network~\cite{skinclassification}.
We remark that our ResNet was originally trained on eight classes, while the RF as binary classifiers.

Nowadays, the detection of melanomas is one of the most researched topics in the oncologic domain~\cite{specialissuemelanoma,specialissuemelanoma1,specialissuemelanoma2}. It is still extremely hard, both for clinicians and computers, to reliably predict a nevus oncologic transformation into a malignant melanoma. 
In order to address this evolutionary issue, future research need to take into account the evolution of the same oncologic data during time.
Our methods and discovery can hopefully help clinicians (or automatic systems) to precisely assess the likelihood of a benign skin lesion to transform into melanoma.

\section{Conclusion}
\label{sec:conclusion}
In this paper we have shown how it is possible to instantiate methodologies of classification and post-hoc explanation in a real case study for skin lesion detection.
In particular, we have proved that, after being customized and trained carefully, \abele{} is able to produce meaningful explanations that can really help practitioners.
The not trivial and time consuming step is the training of the generative model.
\abele{} explanations allowed us to perform two analysis.
The latent space analysis suggests an interesting repartition of image over the latent space, it can hopefully be helpful in separating apart similar classes of skin lesion that are frequently misclassified by humans (benign from malignant).
Also, we conducted a survey involving real user experts and not expert of skin cancer and of healthcare domains.
The survey supports the hypothesis that explanation methods without a consistent validation are not useful.
As future research directions, it will be interesting to apply \abele{} explainer to different diseases and health domain, especially domains where the only meaningful data is the raw image or scan of a particular internal body portion.
Indeed, in skin lesion cancer also the touch plays an important role in doctors' analysis, not just the image.
Also, we would like to extend the user interaction module presented to a real time explanations generator.
This improvement would require a lot of efforts and resources as in the current implementation extracting a single explanation can be time consuming depending on the image.

\section*{Acknowledgment}
This work is partially supported by the European Community H2020 programme under the funding schemes:
G.A. 825619 \emph{AI4EU} ({\texttt{ai4eu.eu}}),
G.A. 952026 \emph{HumanE AI Net} ({\texttt{humane-ai.eu}}), 
G.A. 952215 \emph{TAILOR} ({\texttt{tailor-network.eu}}), and the ERC-2018-ADG G.A. 834756 ``XAI'': Science and technology for the eXplanation of AI decision making''. 

\bibliographystyle{IEEEtran}
\bibliography{biblio.bib}

\end{document}